\documentclass[10pt,twocolumn,letterpaper]{article}

\usepackage{cvpr}              

\usepackage[dvipsnames]{xcolor}

\definecolor{cvprblue}{rgb}{0.21,0.49,0.74}
\usepackage[pagebackref,breaklinks,colorlinks,citecolor=cvprblue]{hyperref}

\def\paperID{110} 
\def\confName{CVPR}
\def\confYear{2024}

\title{Empowering Image Recovery: A Multi-Attention Approach}
\author{Juan Wen$^1$$^2$ \quad Yawei Li$^2$ \quad Chao Zhang$^3$ \quad Weiyan Hou$^1$ \quad Radu Timofte$^2$$^4$ \quad Luc Van Gool$^2$\\ \\
$^{1}$ Zhengzhou University \quad \quad 
$^2$ Computer Vision Lab,ETH Zurich   \quad \quad \\
$^3$  LAN-XEN,Technology,INC. \quad 
$^4$Bayerische Julius-Maximilians-Universität Würzburg
}

\usepackage{stfloats} 
\usepackage{caption} 

\usepackage[utf8]{inputenc}

\usepackage{float}
\usepackage{color}

\usepackage{colortbl}
\usepackage{xcolor}
\usepackage{multirow}
\usepackage{graphicx}
\usepackage{indentfirst}

\begin{document}
\maketitle 
\begin{abstract}

We propose Diverse Restormer (DART), a novel image restoration method that effectively integrates information from various sources (long sequences, local and global regions, feature dimensions, and positional dimensions) to address restoration challenges. While Transformer models have demonstrated excellent performance in image restoration due to their self-attention mechanism, they face limitations in complex scenarios. Leveraging recent advancements in Transformers and various attention mechanisms, our method utilizes customized attention mechanisms to enhance overall performance. DART, our novel network architecture, employs windowed attention to mimic the selective focusing mechanism of human eyes. By dynamically adjusting receptive fields, it optimally captures the fundamental features crucial for image resolution reconstruction. Efficiency and performance balance are achieved through the LongIR attention mechanism for long sequence image restoration. Integration of attention mechanisms across feature and positional dimensions further enhances the recovery of fine details. Evaluation across five restoration tasks consistently positions DART at the forefront. Upon acceptance, we commit to providing publicly accessible code and models to ensure reproducibility and facilitate further research.

\end{abstract}
   
\vspace{-6mm}
\section{Introduction}
\label{sec:intro}
\vspace{-1mm}

\begin{figure}
    \centering
    \scriptsize
    \includegraphics[width=0.85\linewidth]{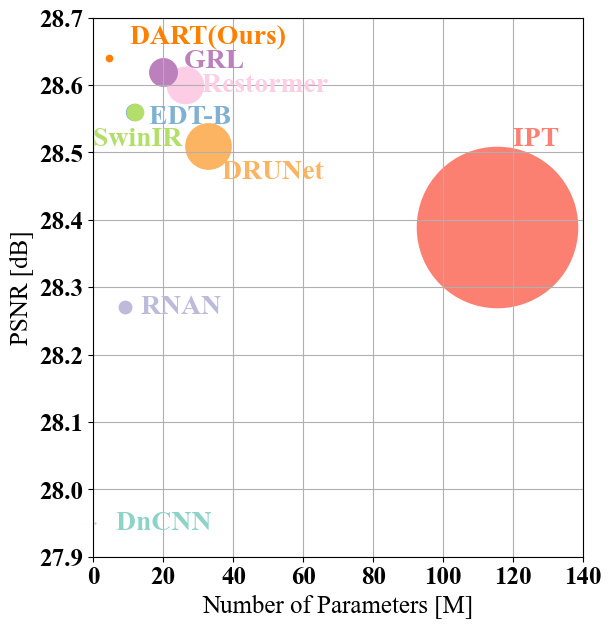}
    \put(-28., 41){}
    \vspace{-4mm}
        \caption{(Color image Denoising CBSD68 dataset) (Noise level: 50)Our DART-B network performs denoising tasks with just 4.5M parameters, achieving the state-of-the-art level for this task. Prior works such as GRL-B ~\cite{li2023efficient} utilized 19.81M parameters, Restormer ~\cite{zamir2022restormer} used 26.13M parameters, and SwinIR~\cite{liang2021swinir} employed 11.75M parameters.}
    \label{fig:teaser}
     \vspace{-5mm}
\end{figure}

Image restoration constitutes a foundational area of research within computer vision and image processing,  
 exerting a pivotal influence across diverse applications. Its primary objective is to enhance image quality, thereby facilitating a more precise and vivid representation of objects and scenes, ultimately enriching visual perception. Specifically, it encompasses the process of improving or recovering the quality and visual fidelity of digital images that have undergone degradation or damage due to factors such as noise, blur, or other forms of distortion.

Transformers have garnered widespread acclaim for their effectiveness in bolstering visual perception and enhancing the performance of various computer vision tasks, including image restoration~\cite{liang2021swinir,liu2021swin}. Spearheaded by Vaswani \emph{et al.}~\cite{vaswani2017attention}Liang \emph{et al.}~\cite{liang2021swinir} and Li \emph{et al.}~\cite{li2023efficient}, underscore the remarkable capabilities of Transformers in tackling challenges associated with image restoration, encompassing tasks like noise reduction and blur correction.

\begin{figure}
    \centering
    \scriptsize
    \includegraphics[width=0.85\linewidth]{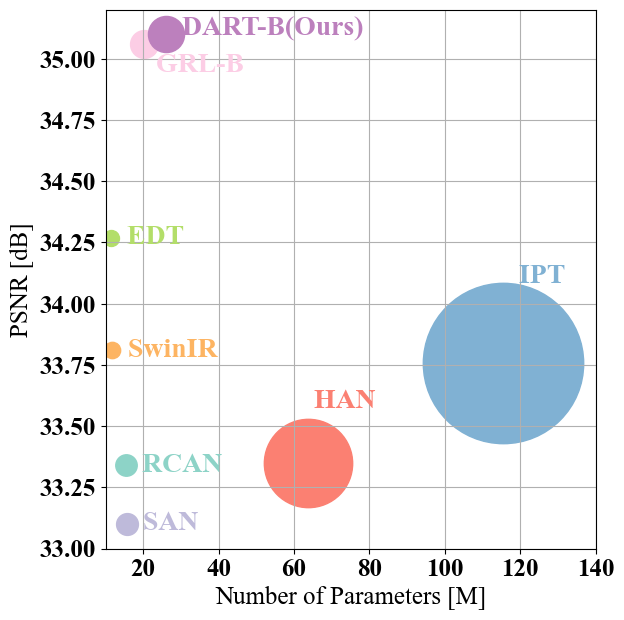}
    \put(-28., 41){}
    \vspace{-5mm}
        \caption{Image SR$\times2$ on Urban100 dataset.}
    \label{fig:teaser}
     \vspace{-6mm}
\end{figure}

While Transformer models excel in capturing complex relationships within data through their self-attention mechanism, they may face challenges in achieving significant gains in certain intricate image restoration tasks. For instance, Restormer has demonstrated substantial improvements across multiple image restoration tasks~\cite{zamir2022restormer}. Introducing a Channel-based self-attention mechanism, Restormer effectively addresses specific challenges in image restoration tasks. However, its complexity increases with the number of channels, rendering it impractical for high-resolution image restoration tasks.

To address the limitations of existing models, we propose a novel multi-attention transformer for image restoration—Diverse Attention Fusion Restoration Transformer (DART), illustrated in Figure 3. DART incorporates several key designs within its attention mechanism module, enabling the model to focus on various aspects of complex patterns and enhance its ability to recover fine details. Extensive experiments unequivocally demonstrate the effectiveness of our proposed Diverse Restormer model, showcasing its prowess in handling complex image restoration tasks.

We utilize the SwinIR network as the infrastructure for our model~\cite{liang2021swinir}. An essential module in our Diverse Attention Fusion Restoration Transformer (DART) model is the Long Sequence Image Restoration (LongIR) module (refer to Figure 3, DART(C)). To address the limitations of SwinIR in handling long sequences due to the quadratic scaling of self-attention operations with sequence length, we propose the LongIR approach. This module employs an attention mechanism that linearly scales with sequence length, enabling it to effortlessly handle thousands of tokens or even longer sequences. We compute the attention for LongIR using $X_w$ (where $X_d$ denotes the input). Subsequently, we combine the window attention of multiple heads with LongIR attention and pass the result to fully connected layers to obtain the merged $X$. Finally, we apply another crucial module in our DART model called feature dimension attention and position dimension attention (refer to Figure 4) to perform additional attention computations on the merged data $X$, yielding the Refined Feature $X$.

The integration of the Long Sequence Image Restoration module facilitates the effective management of sequences through the combination of global and local attention mechanisms. Feature dimension and position dimension attention mechanisms assist neural networks in focusing on specific information within feature maps. Feature dimension attention enhances the representation and utilization of different feature dimensions (feature maps) within the feature map by emphasizing or de-emphasizing certain feature dimensions to improve the network's performance on specific tasks. Conversely, position dimension attention selectively focuses on specific spatial dimensions within images or feature maps, allowing the model to emphasize or de-emphasize certain parts of the input data, thereby enhancing its ability to capture relevant information and improve task performance. This design approach leverages LongIR Attention, Position Dimension Attention, and Feature Dimension Attention, enabling the model to extract information from long sequences, local and global contexts, as well as specific feature dimensions and different Positional Dimension dimensions.

\vspace{-4mm}
\begin{figure*}
    \centering
    \includegraphics[width=0.95\linewidth]{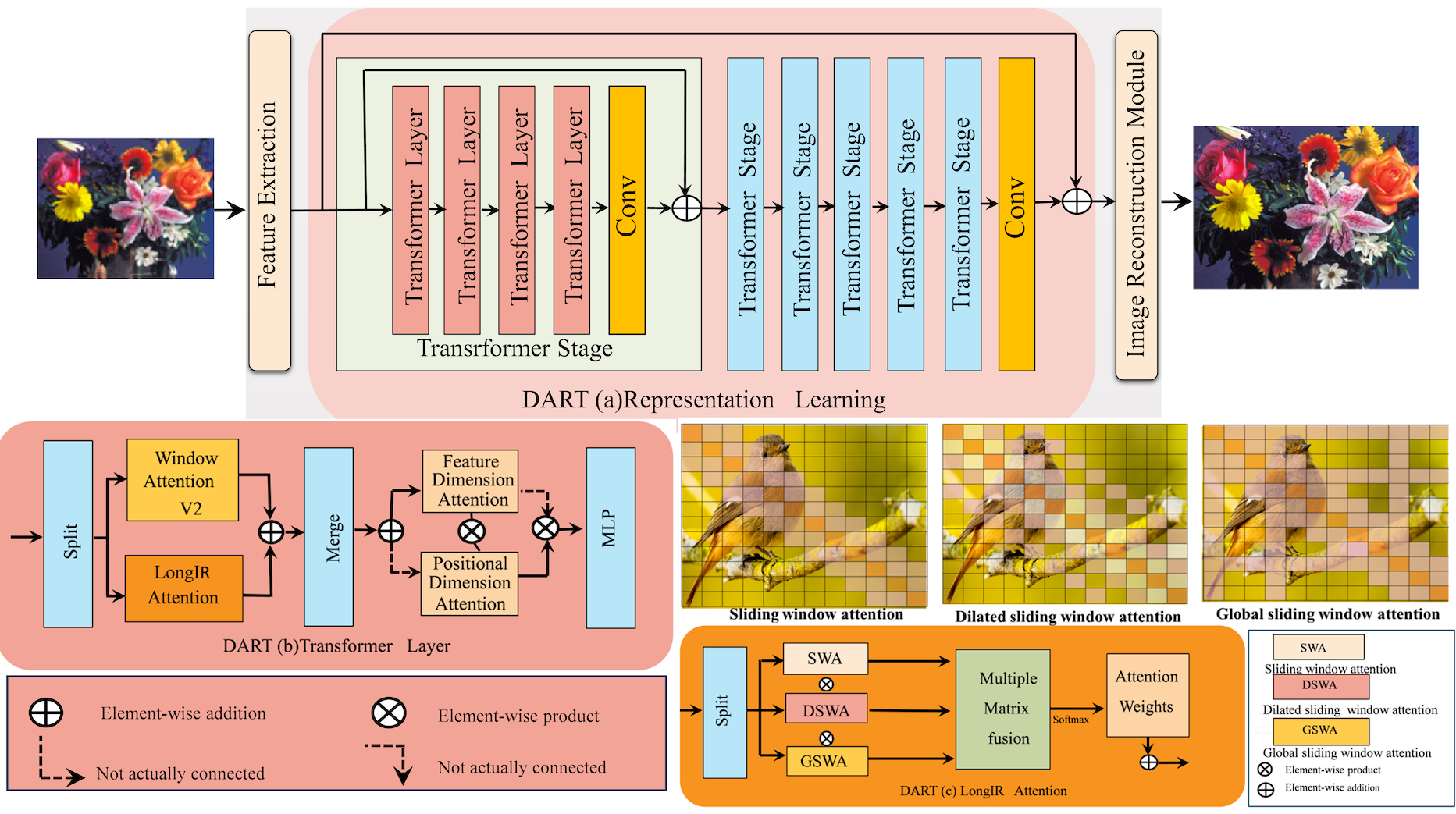}
    \vspace{-4mm}
    \caption{Network Architecture. (a) Illustrates the learning module comprising stages of transformer layers. (b) The Transformer module utilizes LongIR (Sliding Window Attention, Expanded Sliding Window Attention, Global Attention) along with Position Dimension Attention and Feature Dimension Attention mechanisms to extract information from long sequences, local, global, specific feature dimensions, and different Positional Dimension dimensions regions for image restoration. (c) Demonstrates the working mechanism of LongIR Attention.}
    \label{fig:network}
    \vspace{-6mm}
\end{figure*}

\vspace{0.5cm}
In summary, our paper presents three key contributions:

\textbf{Novel Multi-Attention Transformer:} We propose Collaborative Image Restoration with Integrated Attention Mechanisms, a novel multi-attention transformer (DART). DART integrates LongIR along with Position Dimension Attention and Feature Dimension Attention mechanisms, enhancing the model's ability to recover fine details from various complex patterns.

\textbf{Efficient Experimental Design: } To strike a balance between performance and efficiency, we propose an effective experimental design. Our collaborative image restoration model, incorporating multi-attention mechanisms, reduces computational load by decreasing the number of model layers without a significant impact on performance.

\textbf{State-of-the-Art Results:} Our approach achieves state-of-the-art results in image super-resolution, denoising, motion deblurring, and defocus deblurring. Notably, our DART-B network outperforms previous works with only 4.5M parameters, showcasing superiority over GRL-B~\cite{li2023efficient} (19.81M parameters), Restormer~\cite{zamir2022restormer} (26.13M parameters), and SwinIR~\cite{liang2021swinir} (11.75M parameters) in denoising tasks.
\vspace{-2mm}


\vspace{-0.8mm}
\section{Related Works}
\vspace{-1mm}
\label{sec:formatting}
\vspace{-1mm}
\subsection{Image Restoration}
\vspace{-1mm}
In recent years, data-driven CNN architectures~\cite{anwar2020densely, dudhane2022burst, zamir2020learning, yue2020dual, zhang2018image, zhang2020residual} have consistently outperformed traditional restoration methods~\cite{he2010single, kopf2008deep, michaeli2013nonparametric, timofte2013anchored}. Among convolutional designs, U-Net architectures~\cite{zhang2021plug, abuolaim2020defocus, cho2021rethinking, kupyn2019deblurgan, wang2022uformer, yue2020dual, zamir2021multi} have been extensively studied for restoration, thanks to their hierarchical multi-scale representation and computational efficiency. Similarly, skip connection-based approaches have proven effective for restoration, focusing on learning residual signals~\cite{gu2019self, liu2019dual, zamir2020learning, zhang2019residual}. Spatial and channel attention modules have been incorporated to selectively attend to relevant information~\cite{li2018recurrent, zamir2020learning, yue2020dual}. For major design choices in image restoration, readers can refer to NTIRE challenge reports~\cite{abdelhamed2019ntire, abuolaim2021ntire, ignatov2019ntire, nah2021ntire} and recent literature reviews~\cite{anwar2020deep, li2019single, tian2020deep}.

\vspace{-0.1cm}
\subsection{Vision Transformer}
\vspace{-2mm}
The Restormer image restoration model proposed by Zamir et al. exhibits increased computational complexity with the growing number of channels, making it impractical for high-resolution image restoration tasks~\cite{zamir2022restormer}Inspired by the success of pre-training transformer-based models in natural language processing (NLP), researchers introduced a transformer model designed for image restoration in 2021. This model, known as the Image Processing Transformer (IPT)~\cite{chen2021pre}, is noteworthy. However, IPT employs an extensive parameter count (exceeding 115.5M), utilizes large-scale datasets (over 1.1M images), and incorporates multi-task learning for optimal performance.
\vspace{-0.1cm}
In this paper, we introduce the Diverse Restormer (DART) approach, which employs model integration to effectively extract information from long sequences, local and global contexts, specific feature dimensions, and various positional dimension regions. The model aims to strike a balance between efficiency and performance, leveraging multiple attention mechanisms to focus on different aspects of complex patterns, thereby enhancing its ability to recover fine details~\cite{liang2021swinir, liu2021swin, zamir2022restormer}.


\begin{figure*}
    \centering
    \includegraphics[width=0.91\linewidth]{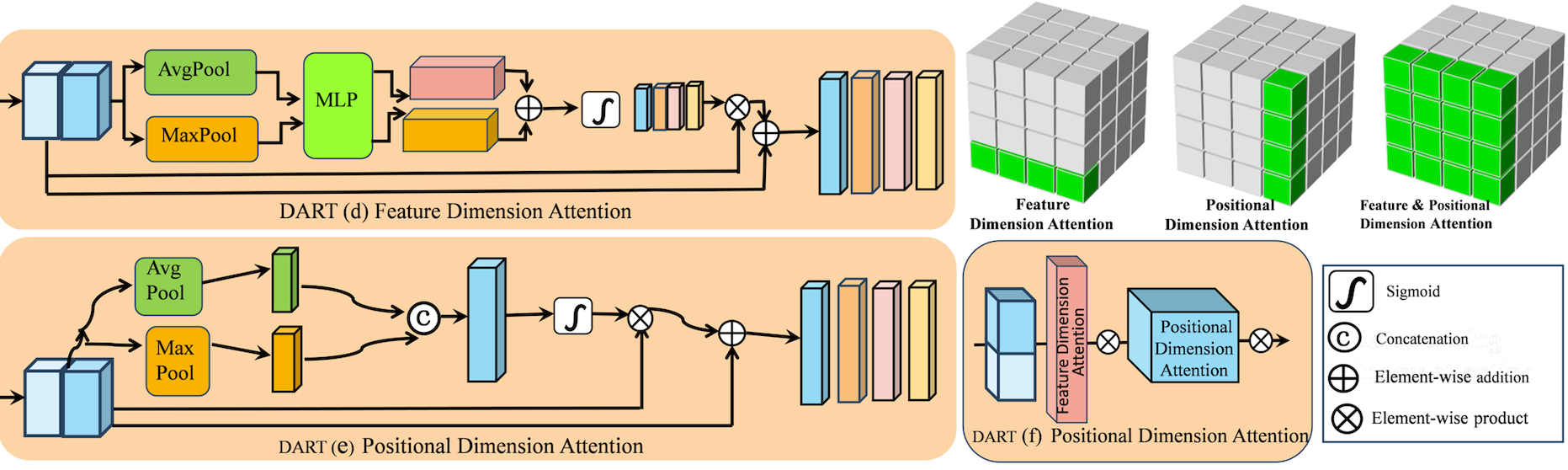}
    \vspace{-4mm}
    \caption{(d), (e), and (f) illustrate the working mechanisms of Position Dimension Attention and Feature Dimension Attention.}
    \label{fig:network}
    \vspace{-6mm}
\end{figure*}

\vspace{-4mm}
\section{Method}
\vspace{-1mm}
In this section, we address the primary research question: Why is collectively extracting information from long sequences, local and global contexts, specific feature dimensions, and various positional dimension regions crucial for image restoration?

\subsection{Motivation}

\vspace{-0.1cm}
The Swin Transformer ~\cite{liu2021swin} has demonstrated excellent performance in image super-resolution. Therefore, we have chosen SwinIR~\cite{liang2021swinir} as the framework for our entire network architecture. We observed that the SwinIR ~\cite{liang2021swinir} model based on the Transformer is unable to handle long sequences due to its self-attention operations, which scale quadratically with sequence length. To address this limitation, we introduced the LongIR (See Figure 3(c)) with an attention mechanism that linearly extends with the sequence length, allowing it to easily handle thousands of tokens or longer information. It can also combine local window attention with task-driven global attention. Furthermore, the transformer model replaces attention matrices with sparse matrices to improve speed.

We reviewed the literature on image restoration over the past three years and found that the Restormer model proposed by Zamir \emph{et al.}~\cite{zamir2022restormer} has performed well in the field. However, its computational complexity grows quadratically with the number of channels, making it impractical for high-resolution image restoration tasks. To address this issue, we adopt feature dimension attention and position dimension attention to assist neural networks in focusing on information within feature maps. Feature dimension attention emphasizes or attenuates specific features to enhance the network's performance on particular tasks. Position dimension attention selectively attends to specific spatial regions within images or feature maps, allowing the model to highlight or downplay certain parts of the input data, thereby enhancing its ability to capture relevant information and improve task performance.

Chen \emph{et al.} proposed a transformer model for image restoration called Image Processing Transformer (IPT)~\cite{chen2021pre}. However, IPT relies on a large number of parameters (exceeding 115.5M parameters), large-scale datasets (over 1.1M images), and multi-task learning for good performance. In order to design a complementary attention mechanism to address different problems in the task, we propose the Diverse Restorer (DART) image restoration model.

\subsection{Network architecture}
Our network architecture, illustrated in Figure 3(a), comprises three main modules: shallow feature extraction, deep feature extraction, and high-quality image reconstruction. The first and last modules rely on CNN, while the intermediate module utilizes SwinTransformer~\cite{liu2021swin} (refer to Figure 3(b)). Transformer's deep feature extraction module combines LongIR with (sliding-window attention, expanding sliding-window attention, global attention) and feature dimension attention mechanism, position dimension attention mechanism.

The crucial layer extracts region-specific details from long sequences, local, global contexts, as well as feature dimension and position dimension attention mechanisms. Initially, it utilizes the window attention module from SwinTransformer V2 to process information~\cite{liu2021swin} and compute the long attention module. Subsequently, it integrates information from multiple multi-head attentions and LongIR (refer to Figure 3(c)), passing it to the fully connected layer for feature fusion. Finally, attention computations are performed using position dimension and feature dimension attention mechanisms (see Figure 4(d, e, f) to obtain processed data from five attention modules.

The core mechanism of LongIR~\cite{beltagy2020longformer} uses three attention windows. 

\vspace{0.5cm}
\textbf{Sliding window attention (Extract local information)}
This attention mechanism can be compared to a convolution kernel, and the size of each convolution kernel is our receptive field. The essence is to set a fixed size window w, which stipulates that each Token in the sequence can only be seen
for $W$ adjacent Tokens, $W$ $\frac{1}{2}$ Tokens can be seen on each side of the left and right sides. In this way, the time complexity changes from the original $O(N\times×N)$ to $O(N\times×W)$, where $W$$<$$N$ . Moreover, we don’t need to worry that this setting cannot establish the information of the entire sequence, because the transformer model structure itself is a layer-by-layer structure. The high-level model has a wider receptive field than the bottom layer, so more information can be seen. so it has the ability to model a global representation that integrates all sequence information. Through this setting, LongIR can achieve a better balance between modeling quality and efficiency.

\textbf{Expanding sliding window attention (Extract long sequence information)}
This attention window is set to make up for the lack of length caused by the previous window. A gap of size d is set between the two Tokens without increasing the time complexity. Then the range of the receptive field can be extended to $d\times×w$.

\textbf{Global attention (Extract global information)}
The whole world here is only part of the whole world. Here we only set certain specific tokens to be able to see all other tokens. For other tokens that are not too important, we still use sliding window attention. Therefore, its attention includes self-attention of windowed local contextual information and global attention activated by terminal tasks. Local attention is used to establish local contextual information representation, and global attention is used to establish a complete sequence representation for prediction. As mentioned earlier, the time and space complexity of existing attention calculation methods are $O(n$²$)$, which makes it difficult to train on long sequences. Therefore, the model uses (attention pattern) to sparse the complete self-attention matrix, which perfectly solves this problem.

\textbf{Feature Dimension Attention Module}  
The Feature Dimension attention is computed as:
\begin{equation}\label{eq:third}
\begin{split}
    \mathbf{M_c}(\mathbf{F})&=\sigma(MLP(AvgPool(\mathbf{F}))+MLP(MaxPool(\mathbf{F})))\\
    &=\sigma( \mathbf{W_1}(\mathbf{W_0}(\mathbf{F^{c}_{avg}}))+
    \mathbf{W_1}(\mathbf{W_0}(\mathbf{F^{c}_{max}}))),
\end{split}
\end{equation}
where \(\sigma\) denotes the sigmoid function, \(\mathbf{W_0}\in \mathbb{R}^{C/r\times C}\), and \(\mathbf{W_1}\in \mathbb{R}^{C\times C/r}\). Note that the MLP weights, \(\mathbf{W_0}\) and \(\mathbf{W_1}\), are shared for both inputs and the ReLU activation function is followed by \(\mathbf{W_0}\).

\begin{table*}[t]
\centering
\caption{\textit{\textbf{Color and grayscale image denoising}} results. 
Model complexity and prediction accuracy are shown for better comparison. }
\label{table:denoising}
\vspace{-3mm}
\setlength{\tabcolsep}{1.5pt}
\scalebox{0.7}{
\begin{tabular}{l | r | c c c | c c c | c c c | c c c || c c c | c c c | c c c }
\toprule[0.1em]
\multirow{3}{*}{\textbf{Method}} & \multirow{3}{*}{\# \textbf{Params} [M]} & \multicolumn{12}{c||}{\textbf{Color}} & \multicolumn{9}{c}{\textbf{Grayscale}} \\ \cline{3-23}
& & \multicolumn{3}{c|}{\textbf{CBSD68}~\cite{martin2001database}} & \multicolumn{3}{c|}{\textbf{Kodak24}~\cite{franzen1999kodak}} & \multicolumn{3}{c|}{\textbf{McMaster}~\cite{zhang2011color}} & \multicolumn{3}{c||}{\textbf{Urban100}~\cite{huang2015single}}  & \multicolumn{3}{c|}{\textbf{Set12}~\cite{zhang2017beyond}} & \multicolumn{3}{c|}{\textbf{BSD68}~\cite{martin2001database}} & \multicolumn{3}{c}{\textbf{Urban100}~\cite{huang2015single}} \\
        &  & $\sigma$$=$$15$ & $\sigma$$=$$25$ & $\sigma$$=$$50$ & $\sigma$$=$$15$ & $\sigma$$=$$25$ & $\sigma$$=$$50$ & $\sigma$$=$$15$ & $\sigma$$=$$25$ & $\sigma$$=$$50$ & $\sigma$$=$$15$ & $\sigma$$=$$25$ & $\sigma$$=$$50$ & $\sigma$$=$$15$ & $\sigma$$=$$25$ & $\sigma$$=$$50$ & $\sigma$$=$$15$ & $\sigma$$=$$25$ & $\sigma$$=$$50$ & $\sigma$$=$$15$ & $\sigma$$=$$25$ & $\sigma$$=$$50$ \\ \midrule
DnCNN~\cite{kiku2016beyond}	&\textcolor{red}{0.56}	&33.90	&31.24	&27.95	&34.60	&32.14	&28.95	&33.45	&31.52	&28.62	&32.98	&30.81	&27.59				&32.86	&30.44	&27.18	&31.73	&29.23	&26.23	&32.64	&29.95	&26.26	\\
RNAN~\cite{zhang2019residual}	&8.96	&-	&-	&28.27	&-	&-	&29.58	&-	&-	&29.72	&-	&-	&29.08				&-	&-	&27.70	&-	&-	&26.48	&-	&-	&27.65	\\
IPT~\cite{chen2021pre}	&115.33	&-	&-	&28.39	&-	&-	&29.64	&-	&-	&29.98	&-	&-	&29.71				&-	&-	&-	&-	&-	&-	&-	&-	&-	\\
EDT-B~\cite{li2021efficient}	&11.48	&34.39	&31.76	&28.56	&35.37	&32.94	&29.87	&{35.61}	&{33.34}	&30.25	&35.22	&{33.07}	&{30.16}				&-	&-	&-	&-	&-	&-	&-	&-	&-	\\
DRUNet~\cite{zhang2021plug}	&32.64	&34.30	&31.69	&28.51	&35.31	&32.89	&29.86	&35.40	&33.14	&30.08	&34.81	&32.60	&29.61				&33.25	&30.94	&27.90	&31.91	&29.48	&26.59	&33.44	&31.11	&27.96	\\
SwinIR~\cite{liang2021swinir}	&11.75	&{34.42}	&31.78	&28.56	&35.34	&32.89	&29.79	&{35.61}	&33.20	&30.22	&35.13	&32.90	&29.82				&33.36	&31.01	&27.91	&\textcolor{blue}{31.97}	&29.50	&26.58	&33.70	&31.30	&27.98	\\
Restormer ~\cite{zamir2022restormer}	&26.13	&34.40	&{31.79}	&{28.60}	&\textcolor{blue}{35.47}	&\textcolor{blue}{33.04}	&\textcolor{blue}{30.01}	&{35.61}	&{33.34}	&{30.30}	&35.13	&32.96	&30.02				&{33.42}	&{31.08}	&\textcolor{blue}{28.00}	&31.96	&{29.52}	&\textcolor{blue}{26.62}	&33.79	&31.46	&{28.29}	\\
GRL-B~\cite{li2023efficient}  &19.81	&\textcolor{blue}{34.45}	&\textcolor{blue}{31.82}	&\textcolor{blue}{28.62}	&{35.43}	&{33.02}	&{29.93}	&\textcolor{blue}{35.73}	&\textcolor{blue}{33.46}	&\textcolor{blue}{30.36}	&\textcolor{blue}{35.54}	&\textcolor{blue}{33.35}	&\textcolor{blue}{30.46}				&\textcolor{blue}{33.47}	&\textcolor{blue}{31.12}	&\textcolor{red}{28.03}	&\textcolor{red}{32.00}	&\textcolor{blue}{29.54}	&{26.60}	&\textcolor{blue}{34.09}	&\textcolor{blue}{31.80}	&\textcolor{blue}{28.59}	\\		
DART-B(ours)  &4.5	&\textcolor{red}{34.49}	&\textcolor{red}{31.85}	&\textcolor{red}{28.64}	&\textcolor{red}{35.49}	&\textcolor{red}{33.06}	&\textcolor{red}{30.05}	&\textcolor{red}{35.79}	&\textcolor{red}{33.52}	&\textcolor{red}{30.39}	&\textcolor{red}{35.56}	&\textcolor{red}{33.39}	&\textcolor{red}{30.50}				&\textcolor{red}{33.50}	&\textcolor{red}{31.15}	&\textcolor{red}{28.03}	&\textcolor{red}{32.00}	&\textcolor{red}{29.56}	&\textcolor{red}{26.63}	&\textcolor{red}{34.11}	&\textcolor{red}{31.82}	&\textcolor{red}{28.61}	\\		
\bottomrule[0.1em]
\end{tabular}}
\vspace{-6mm}
\end{table*}

The input feature map $F$ with dimensions $H \times W \times C$ undergoes global max pooling and global average pooling along the width and height, resulting in two $1 \times 1 \times C$ features. These features are then fed into a shared two-layer neural network (MLP). The first layer has C/r neurons (with r as the reduction rate) and uses the ReLU activation function, while the second layer has C neurons. The output from the MLP undergoes element-wise summation and sigmoid activation, generating the final Feature Dimension attention feature denoted as Mc. Finally, the Feature Dimension attention feature Mc undergoes an element-wise multiplication with the input feature map F, producing the input features required for the Positional Dimension attention module.

\textbf{Positional Dimension Attention Module}   
The Positional Dimension attention is computed as:
\begin{equation}\label{eq:forth}
\begin{split}
    \mathbf{M_s}(\mathbf{F})&=\sigma(f^{7\times 7}([AvgPool(\mathbf{F}); MaxPool(\mathbf{F})]))\\
    &=\sigma(f^{7\times 7}([\mathbf{F^{s}_{avg}}; \mathbf{F^{s}_{max}}])),
\end{split}
\end{equation}
where \(\sigma\) denotes the sigmoid function and \(f^{7\times 7}\) represents a convolution operation with the filter size of $7\times 7$.

The feature map F output by the Feature Dimension attention module is used as the input feature map of this module. First, do a Feature Dimension-based global max pooling and global average pooling to obtain two $H×\times W×\times1$ feature maps, and then perform a concat operation (Feature Dimension splicing) on these two feature maps based on the Feature Dimension. Then, after a $7\times 7$ convolution operation ($7\times 7$ is better than $3\times 3$), the dimension is reduced to 1 Feature Dimension, that is, $H×\times W×\times1$. Then the Positional Dimension attention feature, namely $M$s, is generated through sigmoid. Finally, the feature is multiplied by the input feature of the module to obtain the final generated feature .

\section{Experiments}
\label{sec:modelling_image_hierarchy}

\begin{table*}[t]\scriptsize
\center
\begin{center}
\caption{\textbf{\textit{Classical image SR}} results. Results of lightweight models and accurate models are summarized.}
\vspace{-3mm}
\label{tab:sr_results}
\begin{tabular}{l|c|r|c|c|c|c|c|c|c|c|c|c}
\toprule[0.1em]
\multirow{2}{*}{\textbf{Method}} & \multirow{2}{*}{\textbf{Scale}} & \multirow{2}{*}{\# \textbf{Params} [M]} &  \multicolumn{2}{c|}{\textbf{Set5}~\cite{bevilacqua2012low}} &  \multicolumn{2}{c|}{\textbf{Set14}~\cite{zeyde2012single}} &  \multicolumn{2}{c|}{\textbf{BSD100}~\cite{martin2001database}} &  \multicolumn{2}{c|}{\textbf{Urban100}~\cite{huang2015single}} &  \multicolumn{2}{c}{\textbf{Manga109}~\cite{matsui2017sketch}}  
\\
\cline{4-13}
&  &  & PSNR$\uparrow$ & SSIM$\uparrow$ & PSNR$\uparrow$ & SSIM$\uparrow$ & PSNR$\uparrow$ & SSIM$\uparrow$ & PSNR$\uparrow$ & SSIM$\uparrow$ & PSNR$\uparrow$ & SSIM$\uparrow$ 
\\
\midrule[0.1em]
RCAN~\cite{zhang2018image}&	×2&	15.44&	38.27&	0.9614&	34.12&	0.9216&	32.41&	0.9027&	33.34&	0.9384&	39.44&	0.9786\\
SAN~\cite{dai2019second}&	×2&	15.71&	38.31&	0.9620&	34.07&	0.9213&	32.42&	0.9028&	33.10&	0.9370&	39.32&	0.9792\\
HAN~\cite{niu2020single}&	×2&	63.61&	38.27&	0.9614&	34.16&	0.9217&	32.41&	0.9027&	33.35&	0.9385&	39.46&	0.9785\\
IPT~\cite{chen2021pre}&	×2&	115.48&	38.37&	-&	34.43&	-&	32.48&	-&	33.76&	-&	-&	-\\ \hline
SwinIR~\cite{liang2021swinir}&	×2&	\textcolor{red}{0.88}&	38.14&	0.9611&	33.86&	0.9206&	32.31&	0.9012&	32.76&	0.9340&	39.12&	0.9783\\
SwinIR~\cite{liang2021swinir}&	×2&	11.75&	38.42&	0.9623&	34.46&	0.9250&	32.53&	0.9041&	33.81&	0.9427&	39.92&	0.9797\\  
EDT~\cite{li2021efficient}&	×2&	0.92&	38.23&	0.9615&	33.99&	0.9209&	32.37&	0.9021&	32.98&	0.9362&	39.45&	0.9789\\
EDT~\cite{li2021efficient}&	×2&	11.48&	{38.63}&	{0.9632}&	{34.80}&	0.9273&	{32.62}&	0.9052&	34.27&	0.9456&	{40.37}&	{0.9811}\\ 
GRL-S~\cite{li2023efficient}&	×2&	3.34&	38.37&	{0.9632}&	34.64&	{0.9280}&	32.52&	{0.9069}&	{34.36}&	{0.9463}&	39.84&	0.9793\\
GRL-B~\cite{li2023efficient} &	×2&	20.05&	\textcolor{blue}{38.67}&	\textcolor{blue}{0.9647}&	\textcolor{blue}{35.08}&	\textcolor{blue}{0.9303}&	\textcolor{blue}{32.67}&	\textcolor{blue}{0.9087}&	\textcolor{blue}{35.06}&	\textcolor{blue}{0.9505}&	\textcolor{blue}{40.67}&	\textcolor{blue}{0.9818}\\ 
DART-S(ours) &	×2&	4.70&	38.38&	{0.9632}&	34.66&	{0.9281}&	32.55&	{0.9071}&	{34.38}&	{0.9464}&	39.86&	0.9795\\
DART-B(ours) &	×2&	25.99&	\textcolor{red}{38.69}&	\textcolor{red}{0.9648}&	\textcolor{red}{35.11}&	\textcolor{red}{0.9304}&	\textcolor{red}{32.72}&	\textcolor{red}{0.9089}&	\textcolor{red}{35.10}&	\textcolor{red}{0.9507}&	\textcolor{red}{40.71}&	\textcolor{red}{0.9820}\\
\midrule[0.1em]
RCAN~\cite{zhang2018image}&	×3&	15.44&	34.74&	0.9299&	30.65&	0.8482&	29.32&	0.8111&	29.09&	0.8702&	34.44&	0.9499\\
SAN~\cite{dai2019second}&	×3&	15.71&	34.75& 0.9300& 30.59& 0.8476& 29.33& 0.8112& 28.93& 0.8671& 34.30& 0.9494\\
HAN~\cite{niu2020single}&	×3&	63.61&	34.75& 0.9299& 30.67& 0.8483& 29.32& 0.8110& 29.10& 0.8705& 34.48& 0.9500\\
IPT~\cite{chen2021pre}&	×3&	-&	34.81& -& 30.85& -& 29.38& -& 29.49& -&	-&	-\\ \hline
SwinIR~\cite{liang2021swinir}&	×3&	-&	34.62& 0.9289& 30.54& 0.8463& 29.20& 0.8082& 28.66& 0.8624& 33.98& 0.9478\\  
SwinIR~\cite{liang2021swinir}&	×3&	\textcolor{red}{0.9}&	\textcolor{blue}{34.97}& \textcolor{blue}{0.9318}& \textcolor{blue}{30.93}& 0.8534& \textcolor{blue}{29.46}& \textcolor{blue}{0.8145}& 29.75& 0.8826& 35.12& 0.9537\\
EDT~\cite{li2021efficient}&	×3&	11.48&	\textcolor{blue}{34.97}& 0.9316& 30.89& 0.8527& 29.44& 0.8142& 29.72& 0.8814& 35.13& 0.9534\\ 
DART-S(ours) &	×3&	3.99&	34.76&	{0.9305}&	30.91&	\textcolor{blue}{0.8535}&	29.21&	0.8085&	\textcolor{blue}{29.88}&	\textcolor{blue}{0.8823}&	\textcolor{blue}{35.15}&	\textcolor{blue}{0.9538}\\
DART-B(ours) &	×3&	20.85&	\textcolor{red}{35.10}&	\textcolor{red}{0.9328}&	\textcolor{red}{31.05}&	\textcolor{red}{0.8555}&	\textcolor{red}{29.55}&	\textcolor{red}{0.8163}&	\textcolor{red}{30.22}&	\textcolor{red}{0.8888}&	\textcolor{red}{35.46}&	\textcolor{red}{0.9551}\\
\midrule[0.1em]
RCAN~\cite{zhang2018image}&	×4&	15.59&	32.63&	0.9002&	28.87&	0.7889&	27.77&	0.7436&	26.82&	0.8087&	31.22&	0.9173\\
SAN~\cite{dai2019second}&	×4&	15.86&	32.64&	0.9003&	28.92&	0.7888&	27.78&	0.7436&	26.79&	0.8068&	31.18&	0.9169\\
HAN~\cite{niu2020single}&	×4&	64.20&	32.64&	0.9002&	28.90&	0.7890&	27.80&	0.7442&	26.85&	0.8094&	31.42&	0.9177\\
IPT~\cite{chen2021pre}&	×4&	115.63&	32.64&	-&	29.01&	-&	27.82&	-&	27.26&	-&	-&	-\\ \hline
SwinIR~\cite{liang2021swinir}&	×4&	\textcolor{red}{0.90}&	32.44&	0.8976&	28.77&	0.7858&	27.69&	0.7406&	26.47&	0.7980&	30.92&	0.9151\\
SwinIR~\cite{liang2021swinir}&	×4&	11.90&	32.92&	0.9044&	29.09&	0.7950&	27.92&	0.7489&	27.45&	0.8254&	32.03&	0.9260\\  
EDT~\cite{li2021efficient}&	×4&	0.92&	32.53&	0.8991&	28.88&	0.7882&	27.76&	0.7433&	26.71&	0.8051&	31.35&	0.9180\\
EDT~\cite{li2021efficient}&	×4&	11.63&	{33.06}&	0.9055&	{29.23}&	0.7971&	{27.99}&	0.7510&	27.75&	0.8317&	{32.39}&	\textcolor{blue}{0.9283}\\ 
GRL-S~\cite{li2023efficient}&	×4&	3.49&	32.76&	{0.9058}&	29.10&	{0.8007}&	27.90&	{0.7568}&	{27.90}&	{0.8357}&	32.11&	0.9267\\
GRL-B~\cite{li2023efficient}&	×4&	20.20&	\textcolor{blue}{33.10}&	\textcolor{blue}{0.9094}&	\textcolor{blue}{29.37}&	\textcolor{blue}{0.8058}&	\textcolor{blue}{28.01}&	\textcolor{red}{0.7611}&	\textcolor{blue}{28.53}&	\textcolor{blue}{0.8504}&	\textcolor{blue}{32.77}&	\textcolor{red}{0.9325}\\ 
DART-S(ours) &	×4&	3.96&	32.80&	{0.9060}&	29.14&	{0.8008}&	27.96&	{0.7571}&	{27.95}&	{0.8443}&	32.13&	0.9268\\
DART-B(ours) &	×4&	20.81&	\textcolor{red}{33.12}&	\textcolor{red}{0.9095}&	\textcolor{red}{29.39}&	\textcolor{red}{0.8059}&	\textcolor{red}{28.02}&	\textcolor{blue}{0.7610}& \textcolor{red}{28.55}& \textcolor{red}{0.8505}& \textcolor{red}{32.81}& \textcolor{red}{0.9326}\\	
\bottomrule[0.1em]
\end{tabular}
\end{center}
\vspace{-4mm}
\end{table*}


\begin{figure*}
    \centering
    \includegraphics[width=0.94\linewidth]{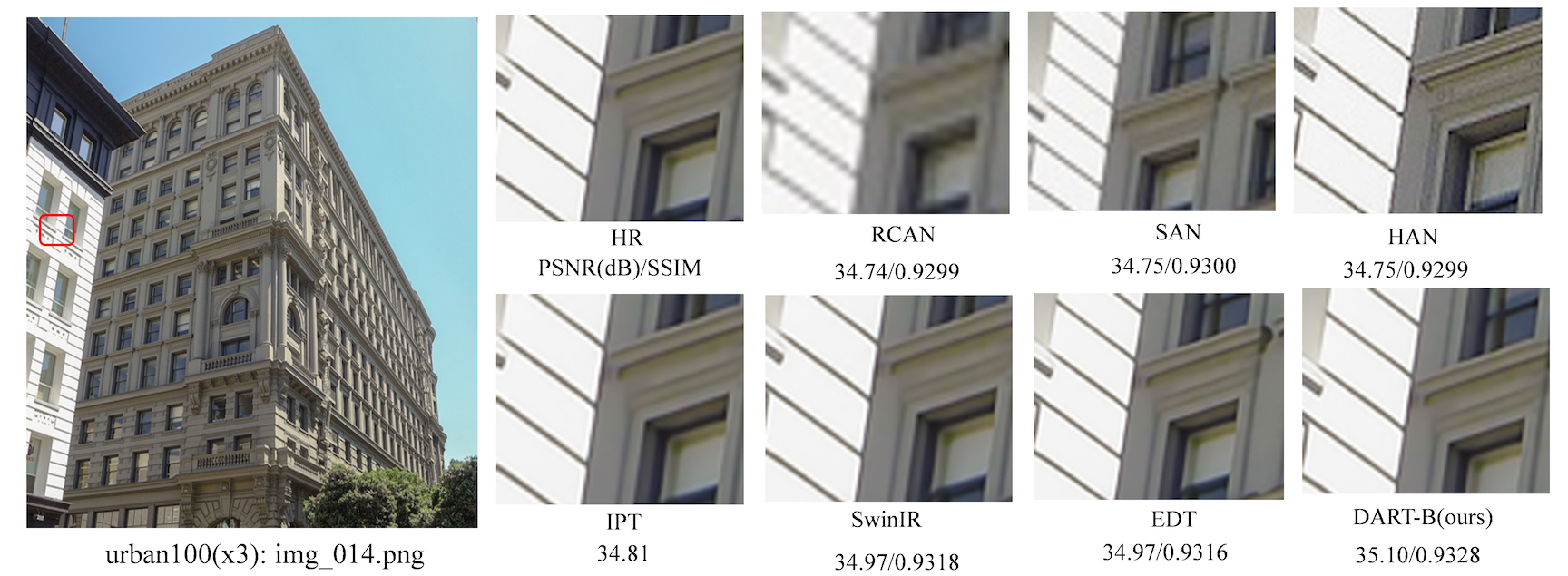}
    \vspace{-4mm}
    \caption{The visual comparison of the DART-B network on x3SR utilizes red bounding boxes to highlight the patch for comparison, in order to better reflect performance differences.}
    \label{fig:network}
    \vspace{-6mm}
\end{figure*}

\begin{figure*}
    \centering
    \includegraphics[width=0.95\linewidth]{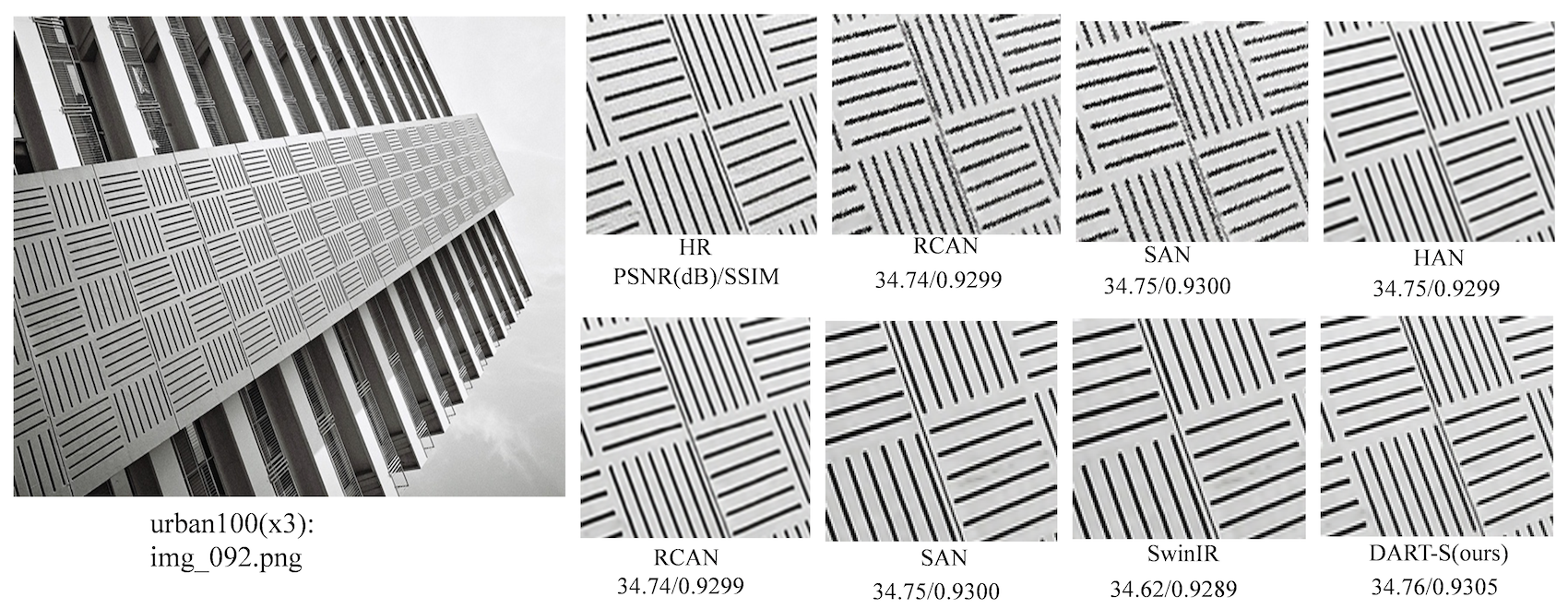}
    \vspace{-4mm}
    \caption{The visual comparison of the DART-S network on x3SR utilizes red bounding boxes to highlight the patch for comparison, in order to better reflect performance differences.}
    \label{fig:network}
    \vspace{-6mm}
\end{figure*}

\begin{table*}[!t]
\begin{center}
\caption{\textit{\textbf\\{Defocus deblurring}} results. \textbf{S:} single-image defocus deblurring. \textbf{D:} dual-pixel defocus deblurring.}
\label{table:defocus_deblurring}  
\vspace{-3mm}
\setlength{\tabcolsep}{1.9pt}
\scalebox{0.7}{
\begin{tabular}{l | c | c | c | c | c | c | c | c | c | c | c | c}
\toprule[0.1em]
\multirow{2}{*}{Method} & \multicolumn{4}{c|}{\textbf{Indoor Scenes}} & \multicolumn{4}{c|}{\textbf{Outdoor Scenes}} & \multicolumn{4}{c}{\textbf{Combined}} \\ \cline{2-13}
	&PSNR$\uparrow$ & SSIM$\uparrow$ & MAE$\downarrow$ & LPIPS$\downarrow$			&PSNR$\uparrow$ & SSIM$\uparrow$ & MAE$\downarrow$ & LPIPS$\downarrow$ &PSNR$\uparrow$ & SSIM$\uparrow$ & MAE$\downarrow$ & LPIPS$\downarrow$		\\ \midrule[0.1em]
EBDB$_S$~\cite{karaali2017edge}	&25.77	&0.772	&0.040	&0.297	&21.25	&0.599	&0.058	&0.373	&23.45	&0.683	&0.049	&0.336	\\
DMENet$_S$~\cite{lee2019deep}	&25.50	&0.788	&0.038	&0.298	&21.43	&0.644	&0.063	&0.397	&23.41	&0.714	&0.051	&0.349	\\
JNB$_S$~\cite{shi2015just}	&26.73	&0.828	&0.031	&0.273	&21.10	&0.608	&0.064	&0.355	&23.84	&0.715	&0.048	&0.315	\\
DPDNet$_S$~\cite{abuolaim2020defocus}	&26.54	&0.816	&0.031	&0.239	&22.25	&0.682	&0.056	&0.313	&24.34	&0.747	&0.044	&0.277	\\
KPAC$_S$~\cite{son2021single}	&27.97	&0.852	&0.026	&0.182	&22.62	&0.701	&0.053	&0.269	&25.22	&0.774	&0.040	&0.227	\\
IFAN$_S$~\cite{lee2021iterative}	&28.11	&0.861	&0.026	&0.179	&22.76	&0.720	&0.052	&0.254	&25.37	&0.789	&0.039	&0.217	\\
Restormer$_S$~\cite{zamir2022restormer}	&{28.87}	&{0.882}	&{0.025}	&{0.145}	&{23.24}	&{0.743}	&{0.050}	&{0.209}	&{25.98}	&{0.811}	&{0.038}	&{0.178}	\\
GRL$_S$-B ~\cite{li2023efficient}	&\textcolor{blue}{29.06}	&\textcolor{blue}{0.886}	&\textcolor{blue}{0.024}	&\textcolor{blue}{0.139}	&\textcolor{blue}{23.45}	&\textcolor{blue}{0.761}	&\textcolor{blue}{0.049}	&\textcolor{blue}{0.196}	&\textcolor{blue}{26.18}	&\textcolor{blue}{0.822}	&\textcolor{blue}{0.037}	&\textcolor{blue}{0.168}	\\ 
DART$_S$-B(ours)	&\textcolor{red}{29.08}	&\textcolor{red}{0.887}	&\textcolor{red}{0.023}	&\textcolor{red}{0.137}	&\textcolor{red}{23.46}	&\textcolor{red}{0.762}	&\textcolor{red}{0.048}	&\textcolor{red}{0.177}	&\textcolor{red}{26.20}	&\textcolor{red}{0.823}	&\textcolor{red}{0.036}	&\textcolor{red}{0.154}	\\ \midrule[0.1em]
DPDNet$_D$~\cite{abuolaim2020defocus}	&27.48	&0.849	&0.029	&0.189	&22.90	&0.726	&0.052	&0.255	&25.13	&0.786	&0.041	&0.223	\\
RDPD$_D$~\cite{lee2021iterative}	&28.10	&0.843	&0.027	&0.210	&22.82	&0.704	&0.053	&0.298	&25.39	&0.772	&0.040	&0.255	\\
Uformer$_D$~\cite{wang2022uformer}	&28.23	&0.860	&0.026	&0.199	&23.10	&0.728	&0.051	&0.285	&25.65	&0.795	&0.039	&0.243	\\
IFAN$_D$~\cite{lee2021iterative}	&28.66	&0.868	&0.025	&0.172	&23.46	&0.743	&0.049	&0.240	&25.99	&0.804	&0.037	&0.207	\\
Restormer$_D$~\cite{zamir2022restormer}	&{29.48}	&{0.895}	&{0.023}	&{0.134}	&{23.97}	&{0.773}	&{0.047}	&{0.175}	&{26.66}	&{0.833}	&{0.035}	&{0.155}	\\
GRL$_D$-B ~\cite{li2023efficient}	&\textcolor{blue}{29.83}	&\textcolor{blue}{0.903}	&\textcolor{blue}{0.022}	&\textcolor{blue}{0.114}	&\textcolor{blue}{24.39}	&\textcolor{blue}{0.795}	&\textcolor{blue}{0.045}	&\textcolor{blue}{0.150}	&\textcolor{blue}{27.04}	&\textcolor{blue}{0.847}	&\textcolor{blue}{0.034}	&\textcolor{blue}{0.133}	\\
DART$_D$-B(ours)	&\textcolor{red}{29.85}	&\textcolor{red}{0.904}	&\textcolor{red}{0.021}	&\textcolor{red}{0.112}	&\textcolor{red}{24.41}	&\textcolor{red}{0.796}	&\textcolor{red}{0.043}	&\textcolor{red}{0.140}	&\textcolor{red}{27.06}	&\textcolor{red}{0.848}	&\textcolor{red}{0.033}	&\textcolor{red}{0.121}	\\
\bottomrule[0.1em]
\end{tabular}}
\end{center}
\vspace{-8mm}
\end{table*}

\begin{table}[!t]
\centering
\caption{\small \textit{\textbf{Single-image motion deblurring}} results. {GoPro} dataset \cite{nah2017deep} is used for training. 
}
\label{table:motion_deblurring}
\vspace{-3mm}
\setlength{\tabcolsep}{1.9pt}
\scalebox{0.69}{
\begin{tabular}{l | c | c | c}
\toprule[0.1em]
 & {\textbf{GoPro}  \cite{nah2017deep}} & {\textbf{HIDE}  \cite{shen2019human}} & Average \\
 \textbf{Method} & PSNR$\uparrow$ / {SSIM$\uparrow$} & PSNR$\uparrow$ / {SSIM$\uparrow$} & PSNR$\uparrow$ / {SSIM$\uparrow$} \\
\midrule[0.1em]
DeblurGAN~\cite{kupyn2018deblurgan}	&28.70 / 0.858		&24.51 / 0.871		&26.61 / 0.865		\\
Nah~\etal~\cite{nah2017deep}	&29.08 / 0.914		&25.73 / 0.874		&27.41 / 0.894		\\
DeblurGAN-v2~\cite{kupyn2019deblurgan}	&29.55 / 0.934		&26.61 / 0.875		&28.08 / 0.905		\\
SRN~\cite{tao2018scale}	&30.26 / 0.934		&28.36 / 0.915		&29.31 / 0.925		\\
Gao \etal \cite{gao2019dynamic}	&30.90 / 0.935		&29.11 / 0.913		&30.01 / 0.924		\\
DBGAN \cite{zhang2020deblurring}	&31.10 / 0.942		&28.94 / 0.915		&30.02 / 0.929		\\
MT-RNN \cite{park2020multi}	&31.15 / 0.945		&29.15 / 0.918		&30.15 / 0.932		\\
DMPHN \cite{zhang2019deep}	&31.20 / 0.940		&29.09 / 0.924		&30.15 / 0.932		\\
Suin \etal \cite{suin2020spatially}	&31.85 / 0.948		&29.98 / 0.930		&30.92 / 0.939		\\
SPAIR~\cite{purohit2021spatially}	&32.06 / 0.953		&30.29 / 0.931		&31.18 / 0.942		\\
MIMO-UNet+~\cite{cho2021rethinking}	&32.45 / 0.957		&29.99 / 0.930		&31.22 / 0.944		\\
IPT~\cite{chen2021pre}	&32.52 / -		&- / -		&- / -		\\
MPRNet~\cite{zamir2021multi}	&32.66 / 0.959		&30.96 / 0.939		&31.81 / 0.949		\\
Restormer~\cite{zamir2022restormer}	&{32.92} / {0.961}		&{31.22} / {0.942}	&{32.07} / {0.952}
		\\
GRL-B~\cite{li2023efficient}	&\textcolor{blue}{33.93} / \textcolor{blue}{0.968}		&\textcolor{blue}{31.65} / \textcolor{blue}{0.947}		&\textcolor{blue}{32.79} / \textcolor{blue}{0.958}	\\
DART-B (ours)	&\textcolor{red}{33.95} / \textcolor{red}{0.969}		&\textcolor{red}{31.70} / \textcolor{red}{0.948}		&\textcolor{red}{32.81} / \textcolor{red}{0.960}	\\
\bottomrule[0.1em]
\end{tabular}}
\vspace{-4mm}
\end{table}

\begin{table}[!t]
\centering
\caption{\small \textit{\textbf{Single-image motion deblurring}} results on {RealBlur} \cite{rim2020real} dataset. The networks are trained and tested on RealBlur dataset. }
\label{table:motion_deblurring_realblur}
\vspace{-3mm}
\setlength{\tabcolsep}{1.9pt}
\scalebox{0.69}{
\begin{tabular}{l | c | c  |c}
\toprule[0.1em]
 & {\textbf{RealBlur-R} \cite{rim2020real}} & {\textbf{RealBlur-J} \cite{rim2020real}} & Average\\
 \textbf{Method} & PSNR$\uparrow$ / {SSIM$\uparrow$} & PSNR$\uparrow$ / {SSIM$\uparrow$} & PSNR$\uparrow$ / {SSIM$\uparrow$}\\
\midrule[0.1em]

DeblurGAN-v2~\cite{kupyn2019deblurgan}	&36.44 / 0.935		&29.69 / 0.870		&33.07 / 0.903	\\
SRN~\cite{tao2018scale}	&38.65 / 0.965		&31.38 / 0.909		&35.02 / 0.937	\\
MPRNet~\cite{zamir2021multi}	&39.31 / 0.972		&31.76 / 0.922		&35.54 / 0.947	\\
MIMO-UNet+~\cite{cho2021rethinking}	&- / -		&32.05 / 0.921		& - / -	\\
MAXIM-3S~\cite{tu2022maxim}	&39.45 / 0.962		&\textcolor{blue}{32.84} / \textcolor{blue}{0.935}		&36.15 / 0.949	\\
BANet~\cite{tsai2022banet}	&39.55 / 0.971		&32.00 / 0.923		&35.78 / 0.947	\\
MSSNet~\cite{kim2022mssnet}	&39.76 / 0.972		&32.10 / 0.928		&35.93 / 0.950	\\
Stripformer~\cite{tsai2022stripformer} & {39.84} / {0.974} & 32.48 / 0.929 &{36.16} / {0.952} \\
GRL-B~\cite{li2023efficient}			&\textcolor{blue}{40.20} / \textcolor{blue}{0.974}		&{32.82} / {0.932}	 &\textcolor{blue}{36.51} / \textcolor{blue}{0.953} \\
DART-B (ours)			&\textcolor{red}{40.23} / \textcolor{red}{0.975}		&\textcolor{red}{32.85} / \textcolor{red}{0.936}	 &\textcolor{red}{36.52} / \textcolor{red}{0.954} \\
\bottomrule[0.1em]
\end{tabular}}
\vspace{-4mm}
\end{table}

\subsection{Experimental Setup}
We assess the performance of our Diverse Restormer (DART) image restoration model across a spectrum of tasks, encompassing:

\textbf{1) Real Image Restoration:}
\begin{itemize}
\item \textbf{Motion Deblurring:} We evaluate our approach on the GoPro and HIDE datasets, comparing against 15 state-of-the-art methods.
\item \textbf{Defocus Deblurring:} Comparative evaluations are conducted on outdoor, indoor, and combined scene datasets, against 14 state-of-the-art methods.
\item \textbf{Real Image Denoising:} We assess performance against 13 state-of-the-art methods using the SIDD and DND datasets.
\end{itemize}

\textbf{2) Synthetic Data-Based Restoration:}
\begin{itemize}
\item \textbf{Denoising:} Evaluation is performed on color (CBSD68, Kodak24, McMaster, Urban100) and grayscale (Set12, BSD68, Urban100) images from multiple datasets, compared against 8 state-of-the-art methods.
\item \textbf{Single Image Super-Resolution (SR):} We examine performance using classical datasets (Set5, Set14, BSD100, Urban100, Manga109) for 2x, 3x, and 4x upscaling comparisons, against 7 advanced methods.
\end{itemize}

We present two model sizes, namely DART-S and DART-B, offering a comprehensive perspective on image restoration ranging from global to local considerations. Additional details on the models, datasets, and results are provided in the supplementary material. The network is trained using the Adam optimizer and L1 loss, with an initial learning rate of $2 \times 10^{-4}$ for both real and synthetic image restoration tasks.

\subsection{Experimental Results}
In this section, we address the second research question by examining the performance of our proposed network across various images. “How does the performance of collaborative image restoration with integrated attention mechanisms compare?"

\textbf{We initially assessed the effectiveness of our proposed Diverse Restormer (DART) network in image restoration tasks using synthetic data.} This included color and grayscale image denoising, classic image super-resolution (SR). Detailed complexities and accuracies are provided in Tables 1 and 2. 1) \textbf{Gaussian Denoising Task:} In this task, our network excels, particularly the DART-B baseline outperforming the state-of-the-art GRL-B method ~\cite{li2023efficient}, despite having only one-fourth of GRL-B's parameters. The more compact DART-S network approaches the performance of the previous state-of-the-art SwinIR ~\cite{liang2021swinir} method, showcasing the effectiveness of our approach compared to the GRL ~\cite{li2023efficient} method. This is particularly evident in focusing on global and local features, as well as long-range and Feature Dimension attention. Additionally, replacing the self-attention method with a long-range attention method in the SwinIR network proves effective in attention modeling. 2) \textbf{Classic Image Super-Resolution (SR):} Results for classic image SR are presented in Table 2, comparing the lightweight SR model with the classic SR model. Similar to Table 1, the conclusions are as follows: DART-S achieves a commendable balance between network complexity and SR image quality in the lightweight network, while DART-B sets a new benchmark for precise image SR, reaching state-of-the-art levels.

\textbf{We extended our network evaluation to real image restoration tasks}, including 3) \textbf{single-image defocus deblurring}, 4) \textbf{dual-pixel defocus deblurring }and 5) \textbf{real image denoising}, 6) \textbf{motion deblurring}. Details are provided in Tables 3, 4, 5, and 6. In Table 4, we present results for single-image motion deblurring on the real image dataset (GoPro ~\cite{nah2017deep}), comparing against the HIDE dataset ~\cite{shen2019human} and in Table 5 the real dataset (RealBlur-R \cite{rim2020real}). Across all three datasets, our method, DART-B, surpasses the state-of-the-art ~\cite{li2023efficient}. Moving to the analysis of denoising performance on real images in Table 6, 
our proposed method outperforms current state-of-the-art techniques.

To further improve the efficiency, we introduce this collaborative image recovery model that integrates multiple attention mechanisms. In another experiment to significantly reduce the model complexity DART-B(s) while maintaining the original DART-B network model, we observed that DART-B(s) has one-third fewer model parameters per SR and a corresponding reduction in training time, but the image restoration results are very close (Table 7). The DART-B(s) and the DART-S(ours) in Table 2 have the network settings and weights are completely different, specifically the experiments of DART-B(s) proved the very strong robustness of our DART network, more details and results are put in the Supplementary Material.

Finally, we conducted a comprehensive ablation study on the Set5 dataset using X2 classical image super-resolution, evaluating the impact of each key component on the proposed model and investigating the effectiveness of three different designs of our proposed model. The results are shown in Table 8. Comparing the LongIR model with the Positional Dimension and Feature Dimension models, we observed improvements of 0.18dB and 0.14dB, respectively, with our DART model.In the LAM (Local Attention Map) experimental study, Figure 7, demonstrate that the proposed DART network exhibits outstanding performance in SR reconstruction by utilizing a wider pixel range.

\begin{table*}[t]
\begin{center}
\caption{\small \small \underline{\textbf{Real image denoising}} on SIDD~\cite{abdelhamed2018high} and DND~\cite{plotz2017benchmarking} datasets. \textcolor{red}{$\ast$} denotes methods using additional training data. \\Our is trained only on the SIDD images and directly tested on DND. }
\label{table:realdenoising}
\vspace{-2.0mm}
\setlength{\tabcolsep}{2.5pt}
\scalebox{0.70}{
\begin{tabular}{c| c| c c c c c c c c c c c c c c  }
\toprule[0.15em]

& { Method}  &CBDNet\textcolor{red}{*}   & RIDNet\textcolor{red}{*}  & AINDNet\textcolor{red}{*}  & VDN & SADNet\textcolor{red}{*} &DANet+\textcolor{red}{*} & CycleISP\textcolor{red}{*} & MIRNet & DeamNet\textcolor{red}{*} & MPRNet & DAGL &  Uformer & Restormer& DART\\
\textbf{Dataset} & &  \cite{guo2019toward} & \cite{anwar2019real} & \cite{kim2020transfer} & \cite{yue2019variational} &  \cite{chang2020spatial}	 & \cite{yue2020dual} &  \cite{zamir2020cycleisp} & \cite{zamir2020learning} & \cite{ren2021adaptive}& \cite{zamir2021multi} & \cite{mou2021dynamic} & \cite{wang2022uformer} & \cite{zamir2022restormer}&(Ours) \\
\midrule[0.15em]
\textbf{SIDD} & PSNR~$\textcolor{black}{\uparrow}$ &  30.78  &  38.71  &  39.08  & 39.28  & 39.46  & 39.47 & 39.52  & 39.72  & 39.47  & 39.71 & 38.94 & {39.77} & \textcolor{blue}{40.02} &\textcolor{red}{40.10}\\
~\cite{abdelhamed2018high}  & SSIM~$\textcolor{black}{\uparrow}$  & 0.801  &  0.951  &  0.954  & 0.956  & 0.957  & 0.957 & 0.957  & 0.959  & 0.957  & 0.958 & 0.953 & {0.959} & \textcolor{blue}{0.960}& \textcolor{red}{0.961}\\
\midrule[0.1em]
\textbf{DND} & PSNR~$\textcolor{black}{\uparrow}$ &   38.06  &  39.26  &  39.37  & 39.38  & 39.59  & 39.58 & 39.56  & 39.88  & 39.63  & 39.80 & 39.77 & {39.96} & \textcolor{blue}{40.03}& \textcolor{red}{40.06} \\
~\cite{plotz2017benchmarking}  & SSIM~$\textcolor{black}{\uparrow}$ &  0.942  &  0.953  &  0.951  & 0.952  & 0.952  & 0.955 & 0.956  & 0.956  & 0.953  & 0.954 & 0.956 & {0.956} & \textcolor{blue}{0.956}&\textcolor{red}{0.957} \\
\bottomrule
\end{tabular}}
\end{center}\vspace{-0.8em}
\end{table*}

\begin{table*}[t]\scriptsize
\center
\begin{center}
\caption{\textbf{\textit{SR-DART-B(s) VS DART-B Classical image SR}} results. By summarizing the performance on Classical image SR of the network model in the original DART-B, without changing any parameters, we directly reduced the model complexity by two orders of magnitude, and compared the performance with the original Classical image SR's metrics.}
\vspace{-3mm}
\label{tab:sr_results}
\begin{tabular}{l|c|r|c|c|c|c|c|c|c|c|c|c}
\toprule[0.1em]
\multirow{2}{*}{\textbf{Method}} & \multirow{2}{*}{\textbf{Scale}} & \multirow{2}{*}{\# \textbf{Params} [M]} &  \multicolumn{2}{c|}{\textbf{Set5}~\cite{bevilacqua2012low}} &  \multicolumn{2}{c|}{\textbf{Set14}~\cite{zeyde2012single}} &  \multicolumn{2}{c|}{\textbf{BSD100}~\cite{martin2001database}} &  \multicolumn{2}{c|}{\textbf{Urban100}~\cite{huang2015single}} &  \multicolumn{2}{c}{\textbf{Manga109}~\cite{matsui2017sketch}}  
\\
\cline{4-13}
&  &  & PSNR$\uparrow$ & SSIM$\uparrow$ & PSNR$\uparrow$ & SSIM$\uparrow$ & PSNR$\uparrow$ & SSIM$\uparrow$ & PSNR$\uparrow$ & SSIM$\uparrow$ & PSNR$\uparrow$ & SSIM$\uparrow$ 
\\
\midrule[0.1em]
DART-B(s) &	×2&	17.52&	\textcolor{blue}{38.66}&	\textcolor{blue}{0.9647}&	\textcolor{blue}{35.06}&	\textcolor{blue}{0.9299}&	\textcolor{blue}{32.69}&	\textcolor{blue}{0.9086}&	\textcolor{blue}{35.07}&	\textcolor{blue}{0.9505}&	\textcolor{blue}{40.65}&	\textcolor{blue}{0.9815}\\
DART-B &	×2&	25.99&	\textcolor{red}{38.69}&	\textcolor{red}{0.9648}&	\textcolor{red}{35.11}&	\textcolor{red}{0.9304}&	\textcolor{red}{32.72}&	\textcolor{red}{0.9089}&	\textcolor{red}{35.10}&	\textcolor{red}{0.9507}&	\textcolor{red}{40.71}&	\textcolor{red}{0.9820}\\
\midrule[0.1em]
DART-B(s) &	×3&	14.16&	\textcolor{blue}{35.01}&	\textcolor{blue}{0.9322}&	\textcolor{blue}{31.00}&	\textcolor{blue}{0.8550}&	\textcolor{blue}{29.50}&	\textcolor{blue}{0.8159}&	\textcolor{blue}{30.18}&	\textcolor{blue}{0.8883}&	\textcolor{blue}{35.40}&	\textcolor{blue}{0.9547}\\
DART-B &	×3&	20.85&	\textcolor{red}{35.10}&	\textcolor{red}{0.9328}&	\textcolor{red}{31.05}&	\textcolor{red}{0.8555}&	\textcolor{red}{29.55}&	\textcolor{red}{0.8163}&	\textcolor{red}{30.22}&	\textcolor{red}{0.8888}&	\textcolor{red}{35.46}&	\textcolor{red}{0.9551}\\
\midrule[0.1em]
DART-B(s) &	×4&	14.12&	\textcolor{blue}{33.03}&	\textcolor{blue}{0.9088}&	\textcolor{blue}{29.32}&	\textcolor{blue}{0.8052}&	\textcolor{blue}{27.98}&	\textcolor{blue}{0.7606}&	\textcolor{blue}{28.51}&	\textcolor{blue}{0.8501}&	\textcolor{blue}{32.78}&	\textcolor{blue}{0.9311}\\
DART-B &	×4&	20.81&	\textcolor{red}{33.12}&	\textcolor{red}{0.9095}&	\textcolor{red}{29.39}&	\textcolor{red}{0.8059}&	\textcolor{red}{28.02}&	\textcolor{red}{0.7610}& \textcolor{red}{28.55}& \textcolor{red}{0.8505}& \textcolor{red}{32.81}& \textcolor{red}{0.9326}\\	
\bottomrule[0.1em]
\end{tabular}
\end{center}
\vspace{-4mm}
\end{table*}
\vspace{-3mm}
 \begin{table}[!t]
\center
\begin{center}

\caption{\textbf{Ablation research} \textit{the key design of the attention module on the 2XSR task of the Set5 data set.}}
\label{ablation}
\setlength{\tabcolsep}{0.8mm}{
\scalebox{0.69}{\begin{tabular}{ccc|cc} 
\hline 
 &LongIR & Feature
,Positional
 &  PSNR & SSIM \\
\hline
& & \checkmark  & 38.27 & 0.9615 \\
& \checkmark &  & 38.31 & 0.9620 \\
& \checkmark & \checkmark  & 38.45 & 0.9625 \\
\hline
\end{tabular}}
}
\vspace{-0.8cm}
\end{center}
\end{table}
\vspace{-2mm}

\begin{figure}
    \centering
    \scriptsize
    \includegraphics[width=0.8\linewidth]{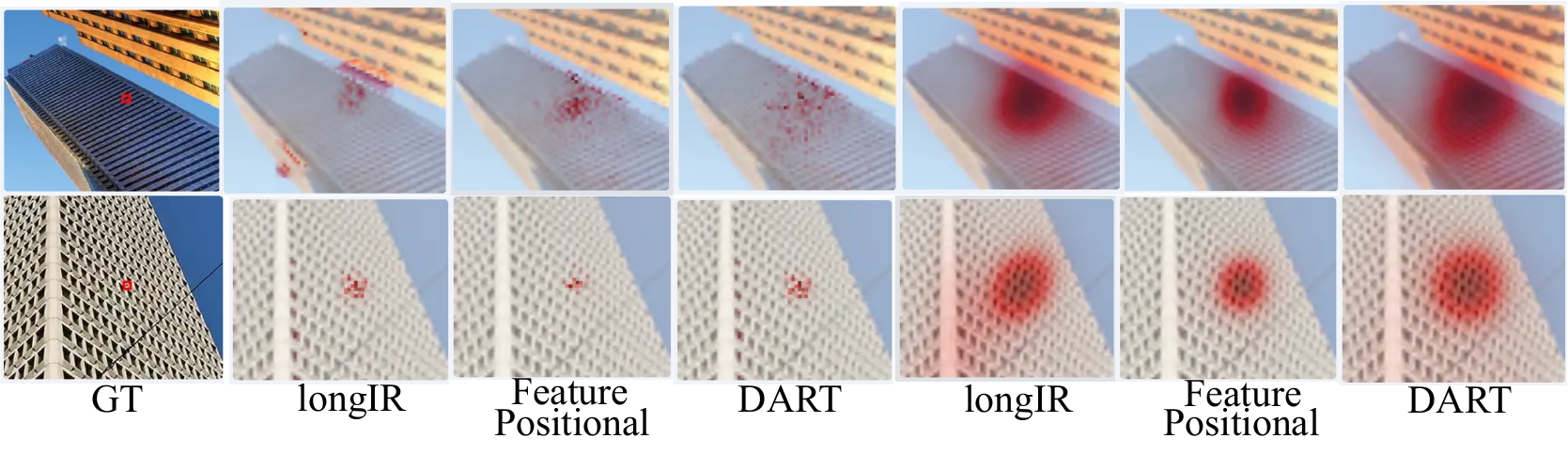}
    \put(-28., 41){}
    \vspace{-4mm}
    \caption{LAM~\cite{gu2021interpreting} different receptive fields in ablation trials. Darker colors indicate greater contributions, larger red areas indicate that a larger range of contextual information was used. Proposed DART network, compared with each key module of the LongIR, feature Dimension, and Positional Dimension networks in the ablation study, demonstrates superior performance in SR reconstruction based on a significantly wider range of pixels.
    }
    \label{fig:teaser}
    \vspace{-5mm}
\end{figure}

\section{Conclusion}

In this paper, we propose DART, a novel and effective image restoration modeling mechanism, a network that integrates the extraction of long sequences, local, global, specific Feature Dimension, and various Positional Dimension information to enable the model to focus on diverse aspects of complex patterns, thereby enhancing its image recovery capabilities to restore fine details. High-resolution images inherently feature intricate patterns and details that demand precise capture, encompassing both Positional Dimension and Feature Dimension (color) information. In the realm of image restoration, it becomes essential to consider features at multiple scales. The design incorporates various attention mechanisms tailored to focus on different scales, facilitating recovery from both local and global contexts. Addressing noise and artifacts in images, which can impede the recovery process, becomes a key aspect. Leveraging diverse attention mechanisms assists in concentrating on noise-free areas while suppressing attention on noisy regions. Based on these insights, we develop the core design of the multi-attention mechanism DART model for image restoration, 
 and our method achieves competitive performance.

\vspace{-3mm}
\newpage                                                


{
    \small
    \bibliographystyle{ieeenat_fullname}
    \bibliography{main}
}


\end{document}